\documentclass[10pt,twocolumn,letterpaper]{article}

\usepackage{cvpr}              

\definecolor{cvprblue}{rgb}{0.21,0.49,0.74}
\usepackage[pagebackref,breaklinks,colorlinks,allcolors=cvprblue]{hyperref}
\usepackage{makecell}
\usepackage{multirow}

\def\paperID{15096} 
\def\confName{CVPR}
\def\confYear{2025}

\title{MSVCOD:A Large-Scale Multi-Scene Dataset for Video \\ Camouflage Object Detection}

\author{Shuyong Gao, Yu'ang Feng, Qishan Wang, Lingyi Hong, Xinyu Zhou, Liu Fei, Yan Wang, Wenqiang Zhang}

\author{
	Shuyong Gao\textsuperscript{\rm 1,2}, Yu'ang Feng\textsuperscript{\rm 1}, Qishan Wang\textsuperscript{\rm 1},\\
	Lingyi Hong\textsuperscript{\rm 1}, Xinyu Zhou\textsuperscript{\rm 1}, Liu Fei\textsuperscript{\rm 2}, Yan Wang\textsuperscript{\rm 1}, Wenqiang Zhang\textsuperscript{\rm 1} \\ 
    \textsuperscript{\rm 1} Fudan University, Shanghai, China \\
    \textsuperscript{\rm 2} Keenon Robotics Co. Ltd, Shanghai, China
    }

\begin{document}


\twocolumn[{
\vspace{-0.3cm}
\maketitle 
\vspace{-0.6cm}
\renewcommand\twocolumn[1][]{#1}%
\vspace{-0.4cm}
\begin{center}
\centering
\includegraphics[width=\textwidth]{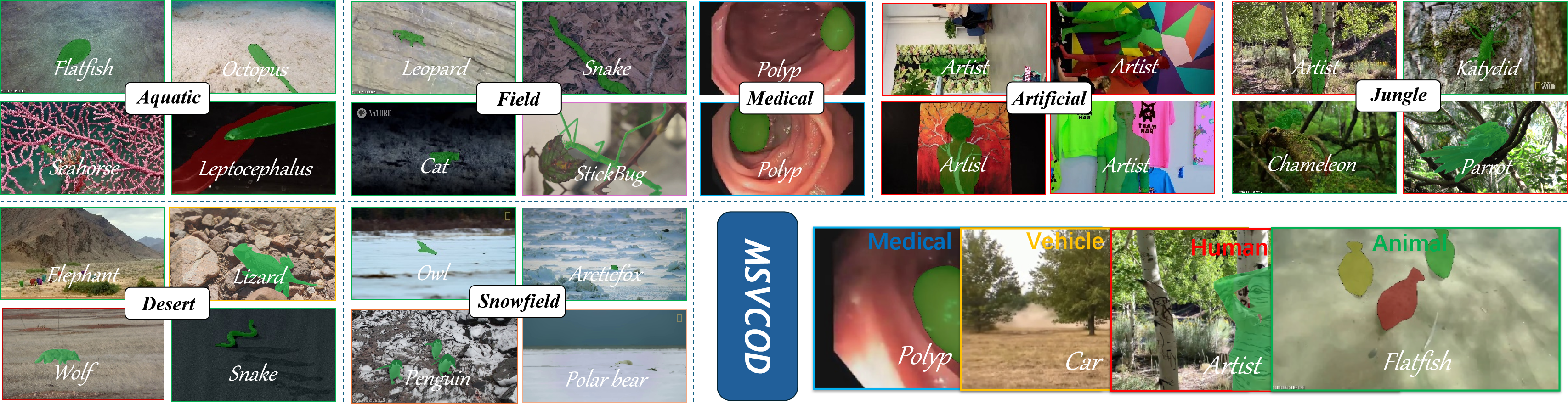}
\vspace{-0.7cm}
\captionof{figure}{An overview of MSVCOD composed of video frames from seven scenarios and four types of object.}\label{fig:fig1}
\vspace{-0.1cm}
\end{center}}]

\begin{abstract}
  Video Camouflaged Object Detection (VCOD) is a challenging task which aims to identify objects that seamlessly concealed within the background in videos. The dynamic properties of video enable detection of camouflaged objects through motion cues or varied perspectives. Previous VCOD datasets primarily contain animal objects, limiting the scope of research to wildlife scenarios. However, the applications of VCOD extend beyond wildlife and have significant implications in industry, art, and medical fields. Addressing this problem, we construct a new large-scale multi-domain VCOD dataset MSVCOD. To achieve high-quality annotations, we design a semi-automatic iterative annotation pipeline that reduces costs while maintaining annotation accuracy. Our MSVCOD is the largest VCOD dataset to date, introducing multiple object categories including human, animal, medical, and vehicle objects for the first time, while also expanding background diversity across various environments. This expanded scope increases the practical applicability of the VCOD task in camouflaged object detection. Alongside this dataset, we introduce a one-steam video camouflage object detection model that performs both feature extraction and information fusion without additional motion feature fusion modules. Our framework achieves state-of-the-art results on the existing VCOD animal dataset and the proposed MSVCOD. The dataset and code will be made publicly available.
\end{abstract}

\section{Introduction}

\begin{figure*}
    \centering
    \includegraphics[width=1\linewidth]{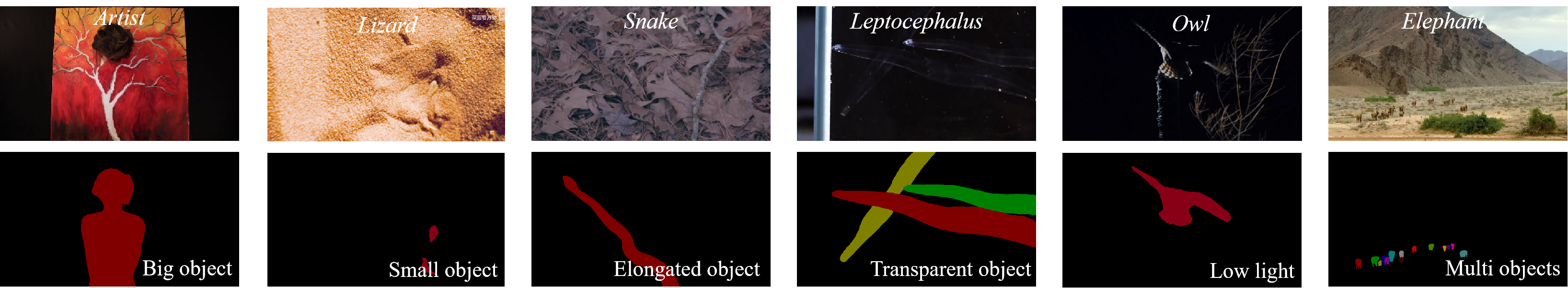}
    \caption{Detail illustration of the data collection and annotation pipline.}
    \label{fig:big_small}
    \vspace{-0.4cm}
\end{figure*}

Camouflaged Object Detection (COD) \cite{fan2020camouflaged} focuses on identifying and segmenting the hidden objects that closely resemble their backgrounds. Those camouflaged objects exhibit intricate visual patterns, edge breaking, texture similarities, and color matching, making them blend seamlessly into their surroundings and challenging to detect compared to traditional object detection methods \cite{Skelhorn2016-CODcognition, Zhao2019-TraditionalOD}. COD has applications in diverse fields such as medical image segmentation \cite{fan2020inf,FanDP2020-Pranet}, enemy detection in the search-and-rescue, art \cite{Ge2018-Art,chu2010camouflage}, industry \cite{Zeng2022-DefectDetection}, scientific research \cite{Perez2012-SpeciesDiscovery}, and agriculture \cite{Rustia2020-Agriculture}. Given that motion and multi-angle visual information can help to detect camouflaged objects, Video Camouflage Object Detection (VCOD) was developed. VCOD, a sub-domain of COD, focuses on segmenting out camouflage targets in the videos by utilizing motion and multi-angle visual cues.

Although video can effectively reveal camouflage, VCOD datasets are relatively scarce due to the limited amount of video camouflage data and the time-consuming, labor-intensive nature of manual labeling. In 2016, Pia Bideau and Erik Learned-Miller introduced the first VCOD dataset, the Camouflaged Animal Dataset (CAD) \cite{bideau2016s}, which includes 9 short clips of camouflaged animals from YouTube videos and provides 191 frames with manual ground truth mask annotations. In 2020, Hala Lamdouar \emph {et al.} proposed the first large-scale VCOD dataset, Moving Camouflaged Animals (MoCA) \cite{lamdouar2020betrayed}, which consists of 141 video clips featuring 67 animals and provides bounding box annotations. On this basis, Cheng \emph {et al.} reorganized and annotated MoCA, resulting in 87 video sequences with ground truth mask annotations every 5 frames, totaling 5,750 annotations. This reorganized dataset, called MoCA-Mask \cite{cheng2022implicit}, is currently the only large-scale dataset in the field.Based on this, Cheng \emph {et al.} reorganized and annotated MoCA, creating MoCA-Mask \cite{cheng2022implicit}, a large-scale dataset with 87 video sequences and 5,750 ground truth mask annotations, which is currently the large-scale dataset in the field.

\begin{table}[ht]
  \centering
  \Large
    \caption{Comparison of MVCOD with other video camouflage object detection dataset benchmark.(Img. = Number of frames in the dataset; Obj. = Types of object; BBox. = Bounding box level; Pix. = Pixel level; Ins. = Instance level; Cate. = Category; Spi. means explicitly splitting the Training and Testing Set)}
  \label{tab_comparison}
  \resizebox{0.47\textwidth}{!}{
  \begin{tabular}{lccc ccccc}
    \toprule
    Dataset      & Clips &Img. &Obj. &BBox. &Pix. &Ins. &Cate. &Spi\\
    \midrule
    CAD$_{2016}$  & 9 &191  &1 & &\checkmark & & &\\
    MoCA$_{2020}$  & 141 &7617  &1 &\checkmark & & & &\\
    MoCA-Mask$_{2022}$  & 87 &5750 &1 &\checkmark &\checkmark & &\checkmark & \checkmark \\
    \midrule
    Ours  &162 &9486 &4 &\checkmark &\checkmark &\checkmark &\checkmark &\checkmark  \\
    \bottomrule

  \end{tabular}
  }
\vspace{-0.3cm}
\end{table}

Current video camouflage datasets primarily focus on animal scenes, overlooking the diverse camouflage modalities human society. This limitation restricts the broader applicability of VCOD, particularly in fields such as medicine, search-and-rescue, and art. The lack of diverse datasets results in researchers not having enough data to train and test models on benchmark datasets. To enhance model generalization across scenes and objects, existing VCOD models \cite{lamdouar2020betrayed, zhang2024explicit} are often pretrained on static image datasets \cite{fan2020camouflaged}. To address this gap, we built a novel, large-scale multi-scene VCOD benchmark dataset. It includes 162 video clips across four object categories (human, animal, medical, and vehicle) and seven scenarios (aquatic, field, medical, art, jungle, desert, snowfield). Our dataset provides 6 frames of ground truth mask annotations per second, totaling 9,486 frame annotations, making it the largest VCOD dataset to date. Table \ref{tab_comparison} and Figure \ref{fig:statistic graphs} show the characteristics of our dataset.


Some static image-based methods detect camouflaged objects from still images by first localizing and then refining \cite{FanDP2020-Pranet, fan2020camouflaged, FanDP2021-ConcealedOD}, sometimes in combination with multitasking \cite{zhai2021MGL, Lv2021-RankNet, he2023camouflaged}. Other models \cite{pang2022zoom, jia2022-CODSegMar, Xing2023-SARNet} achieve significant performance gains by incorporating simple images or feature amplification. However, these models focus solely on still images and cannot take advantage of motion information in videos. To address this, some models \cite{lamdouar2020betrayed, zhang2024explicit} explicitly integrate optical flow information for camouflaged object detection, drawing inspiration from the neighboring field of video salient object detection \cite{lan2022siamese, gao2022weakly, ji2021full}. SLT-Net \cite{cheng2022implicit} extracts image features from consecutive frames separately and then fuses short-term and long-term features to detect camouflaged objects.

The VCOD models discussed above all follow a two-stream architecture, where feature extraction and information fusion are performed separately. This approach is computationally expensive and often leads to poor performance due to the difficulty of adaptively extracting exploitable features. In contrast, inspired by video object tracking and video object segmentation \cite{cui2022mixformer, hong2023simulflow}, and in conjunction with MSVCOD, we propose a one-stream VCOD model. This model simultaneously extracts image features and motion information, eliminating the need for optical flow as input. At the decoding layer, we design a simple, fully-connected, UNet-like decoder that relies on linear adapter layer, without any unnecessary complexity, achieving state-of-the-art performance. Our main contributions are as follows:

\begin{itemize}

\item  We design a semi-automatic iterative annotation pipeline and construct a novel, large-scale multi-scene video camouflage object detection dataset, MSVCOD. The dataset consists of 162 clips and 9,486 frames, covering 7 scenes across 4 major categories, and introduces a wide range of non-wildlife targets for the first time. It provides annotations at the box, mask, instance, and category levels.

\item We develop a simple, one-stream camouflage object detection model equipped with fully-connected UNet-like decoder, enabling simultaneous extraction of image features and fusion of motion features


\item Extensive experiments demonstrate that our proposed dataset enhances model performance and improves generalization across multiple scenarios. Additionally, numerous experiments show that our model significantly outperforms previous VCOD models.

\end{itemize}




\section{Related Work}

\subsection{Dataset for Camouflage Object Detection}

The rapid development of camouflage object detection in recent years is partly due to advancements in deep learning algorithms and the emergence of large-scale camouflage object detection (COD) datasets, which form the foundation for training and benchmarking COD models. Datasets for static camouflage object detection include CAMO \cite{camo}, CHAMELEON \cite{chameleon}, COD10K \cite{fan2020camouflaged}, and NC4K \cite{Lv2021-RankNet}. The CAMO dataset \cite{camo} contains 1,000 training images and 250 test images, covering various challenging scenarios, including camouflaged animals and artificial camouflage. CHAMELEON \cite{chameleon} consists of 76 images, focusing on camouflaged animals. COD10K \cite{fan2020camouflaged}, the largest camouflage object dataset, includes around 10000 images and is frequently used for pre-training VCOD models.

For VCOD, the CAD dataset \cite{bideau2016s} is a small dataset, containing 9 short sequences from YouTube with hand-labeled ground truth masks every 5 frames. The original Moving Camouflaged Animals (MoCA) dataset \cite{lamdouar2020betrayed} includes 37K frames from 141 YouTube videos at 720 × 1280 resolution and 24 fps, featuring 67 animal species in natural scenes (though not all are camouflaged). MoCA provides ground truth as bounding boxes, making segmentation evaluation difficult. Cheng \emph {et al.} \cite{cheng2022implicit} reorganized MoCA into MoCA-Mask, establishing a benchmark with more comprehensive evaluation criteria. However, all existing VCOD datasets are animal-specific. In contrast, we propose the MSVCOD dataset, which covers multiple object types and scenarios.

\subsection{Image-based Camouflage Object Detection}

Image-based COD aims to identify camouflaged objects in still images. Early methods \cite{pan2011study, liu2012foreground} relied on hand-designed features to distinguish camouflaged objects from their background. With the development of deep learning and large-scale COD datasets \cite{camo, chameleon, fan2020camouflaged}, the field has advanced rapidly. Some methods \cite{FanDP2020-Pranet, fan2020camouflaged, FanDP2021-ConcealedOD} use a coarse-to-fine approach to progressively identify camouflaged objects. To further enhance performance, some studies \cite{zhai2021MGL, Lv2021-RankNet, he2023camouflaged} integrate auxiliary tasks within a joint learning framework. Additionally, some works \cite{pang2022zoom, jia2022-CODSegMar, Xing2023-SARNet} explore image or feature amplification to improve camouflage recognition. For example, ZoomNet \cite{pang2022zoom} employs a zoom-in-and-out technique to process appearance features across three different scales. Other methods \cite{lin2023frequency, sun2025frequency, zhang2024frequency} attempt to segment camouflaged objects through frequency analysis. However, since these models are designed for still images, they cannot utilize motion information, which limits their performance in video camouflage object detection tasks.

\subsection{Video Camouflage Object Detection}

Video camouflage object detection is a recently developed research field. Similar to other video segmentation tasks (\emph {e.g.}, VSOD \cite{lan2022siamese, gao2022weakly, ji2021full} and VOS \cite{hong2023simulflow, miao2024region, seong2022video, yang2022decoupling}), motion cues are considered an effective means to break the camouflage of objects. Bideau \emph {et al.} \cite{bideau2016s} use various motion models derived from dense optical flow to detect camouflaged objects. Lamdouar \emph {et al.} \cite{lamdouar2020betrayed}, also using optical flow, propose a network for video registration and segmentation that utilizes optical flow and difference images to detect camouflaged objects. However, their model can only output detection results at the bounding box level due to dataset limitations.

While optical flow \cite{deng2023explicit, teed2020raft, dosovitskiy2015flownet, sun2018pwc} provides motion information, it can be problematic when the object is stationary or only the camera is moving. Additionally, the computational cost of optical flow is high. Cheng \emph {et al.} \cite{cheng2022implicit} propose a two-stage model that implicitly incorporates motion information over long and short time intervals and predicts pixel-level masks. Further, explicitly handling motion cues, Zhang \emph {et al.} \cite{zhang2024explicit} introduce a two-stream model to estimate optical flow and segment camouflaged objects. In contrast to these approaches, and with the support of the MSVCOD dataset, We propose a one-stream camouflage object detection model that does not require explicitly computing optical flow as input.

\section{MSVCOD Dataset}

\begin{figure*}
    \centering
    \includegraphics[width=0.8\linewidth]{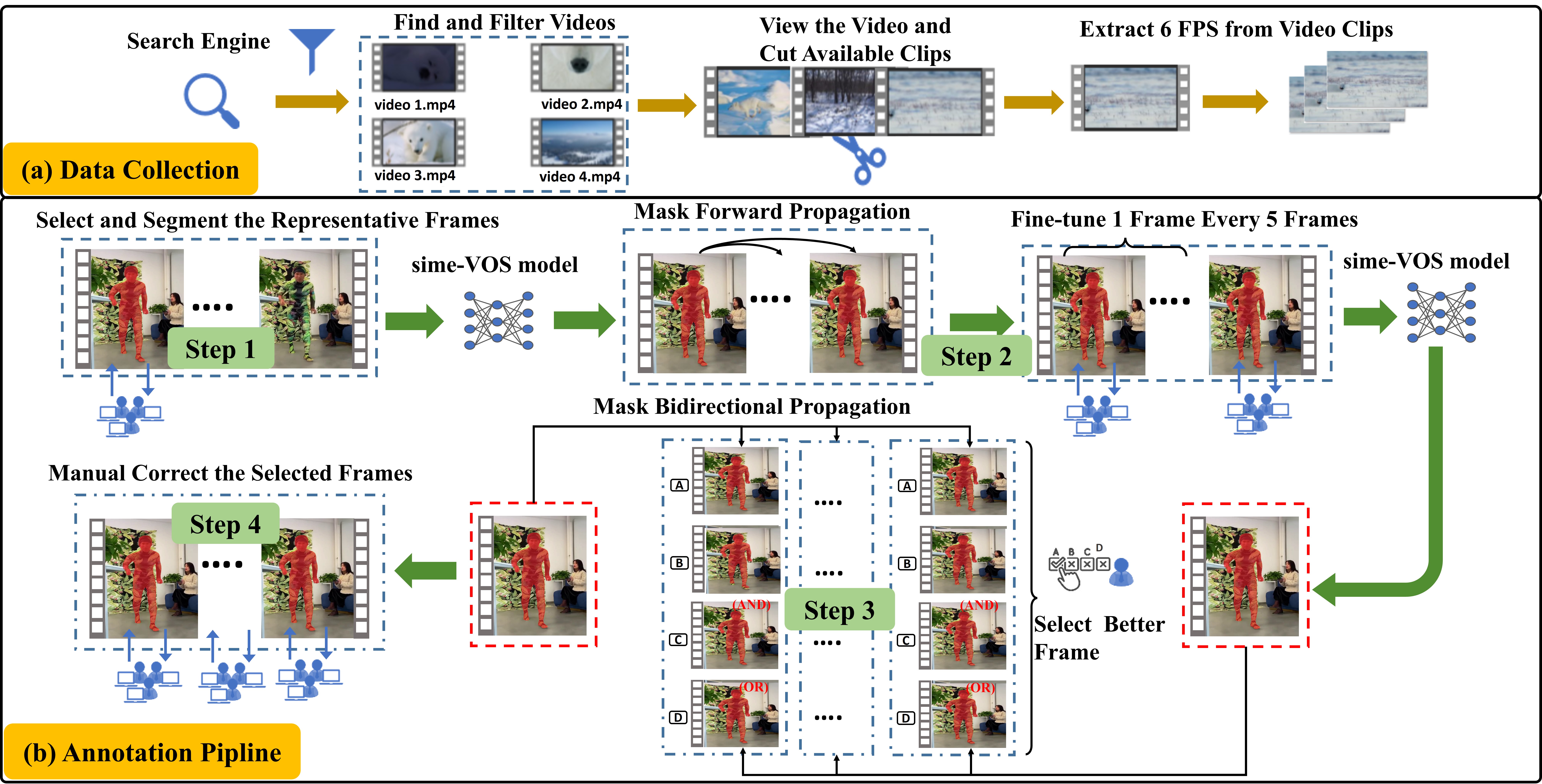}
    \caption{Detail illustration of the data collection and annotation pipline.}
    \label{fig:annotation pipline}

\vspace{-0.3cm}
\end{figure*}

\subsection{Dataset Construction}           

\textbf{Dataset Design and Data Collection.} To address the limitation that current VCOD datasets focus primarily on animal camouflage, with fewer scenes and less data, MSVCOD aims to establish a novel VCOD dataset for training and evaluating robust VCOD models. Our primary focus is on camouflaged object detection in challenging and complex scenes that feature a variety of camouflage objects. To ensure the creation of a high-quality VCOD dataset, we set several guidelines during the data collection and construction process. These requirements are summarized as follows:

\textbf{R1:} Camouflage objects in the videos should be well-camouflaged. Specifically, the objects should be similar to the background and not be easily recognizable by annotators at first glance. The objects can be one or multiple objects.

\textbf{R2:} The collected videos should cover a variety of scenes, including underwater, land, desert, jungle, and other diverse environments.

\textbf{R3:} The dataset should contain various types of targets, including animals, humans, and vehicles.

\textbf{R4:} As shown in Figure \ref{fig:big_small}, targets in the dataset should vary in size and shape, including large, small, elongated targets, or those with gradually changing scales.

\textbf{R5:} The dataset should include various motion patterns, such as camera motion, object motion, or simultaneous motion of both the object and the camera.

\textbf{R6:} The dataset should be large-scale and densely annotated with high-quality data. The scale and quality of the dataset will ensure its long-term utility. Therefore, MSVCOD includes a substantial number of video clips. Given the time-consuming nature of densely annotating high-quality camouflage images, we have developed a semi-automatic iterative annotation method. The data collection process is illustrated in Figure \ref{fig:annotation pipline} (a).



To collect the data meeting the above requirements, we searched for 227 videos containing camouflaged objects and selected seven scenarios with four types of targets: animal, human, medical, and vehicle. The long-duration videos were divided into multiple clips, as in previous work \cite{wang2022ferv39k,ding2023mose}, and the same object potentially appeared in multiple clips. Seven annotators participated in the task, with three experts in camouflage object detection selecting the relevant video clips based on their experience.

\begin{table}[h!]
\centering
\caption{Statistical table of three typical movement patterns.}
\begin{tabular}{lccc}
\toprule
\textbf{Name} & \textbf{Train} & \textbf{Test} & \textbf{Total} \\
\midrule
Object Motion         & 53 & 20 & 73 \\
Camera Motion         & 22 & 7  & 29 \\
Simultaneous Motion   & 46 & 14 & 60 \\
\bottomrule
\end{tabular}
\label{tab:motion patterns}
\end{table}

In the videos, if camouflaged objects become visible due to changes in the background or the objects themselves, we segment and exclude the corresponding clips, as referencing visible parts would reduce the level of camouflage. Finally, considering both video quality and segmentation difficulty, we selected 162 video clips to form MSVCOD.

To comply with \textbf{R4}, We manually selected video clips that contain targets with scale variations and size transformations. For example, objects occupying a larger portion of the frame include artist and octopus; smaller objects include humans in outdoor environments and various insects; and elongated objects include snake and eel. The scatterplot in Figure \ref{fig:msvcod_vs_moca} visually demonstrates the large scale distribution of our dataset. \textbf{R5} requires a variety of motion patterns. Here are some examples. \textbf{Object motion}: In artificial environments, the camera remains stationary while the object moves. \textbf{Camera motion}: In underwater environments, it is sometimes challenging to stabilize the filming equipment. \textbf{Simultaneous motion}: When filming wildlife in the field, the camera tracks the animal’s movement while it moves.
More visual examples are provided in the supplementary material. Additionally, we performed statistics on the video clips across different motion models, as shown in table \ref{tab:motion patterns}.

\textbf{Annotation Method.} To obtain accurate pixel-level annotations for video clips, a purely manual annotation method is feasible when the dataset is small or the objects are easy to annotate. However, with 9.5K frames in this dataset, manually annotating such a large number of camouflage images frame by frame is costly and time-consuming. To address this, we designed a semi-automatic iterative annotation pipeline to progressively improve annotation quality, as shown in Figure \ref{fig:annotation pipline} (b).

\textbf{Step 1: Selection and Segmentation of Representative Frames.} In the automatic annotation process, we utilize semi-supervised video object segmentation algorithms (SwinB-DeAOT-L \cite{yang2022decoupling}) to generate pseudo-labels. The semi-VOS model uses the first frame and its corresponding mask as a reference to identify objects in subsequent frames. Unlike VCOD, where camouflaged objects may not appear in the first frame, the semi-VOS model assumes that objects are already segmented in the first frame. To adapt VOS models for VCOD, we manually select representative frames from each video clip to serve as the reference frame. If the camouflaged object appears in the first frame, we use it as the reference. If the camouflaged object does not appear in the first frame, we choose the frame closest to the first frame that contains the camouflage object as the reference.

\textbf{Step 2: Mask Forward Propagation and 1 FPS Manual Correction.} To automatically generate pseudo-labels based on the manual annotations, we use the VOS model (SwinB-DeAOT-L \cite{yang2022decoupling}) for pseudo-label generation. All input frames are resized to 512 pixels. The VOS model then outputs pseudo-labels for the unannotated frames in each video clip. Following this, we manually correct one frame every six frames (corresponding to our sampling rate) by reviewing and refining the generated pseudo-labels.

\textbf{Step 3: Mask Bidirectional Propagation.} Using the manually corrected mask as the reference frame, we apply the DeATO model \cite{yang2022decoupling} in both the forward and backward directions. As a result, each intermediate frame receives two pseudo-labels. These two pseudo-labels are then combined using the AND and OR operations, generating a total of four pseudo-labels, as shown in Figure \ref{fig:annotation pipline} (b) Step 3. Finally, the pseudo-labels are converted into polygon masks. Instead of selecting the consistency region as the pseudo-label, we allow the annotator to choose the most appropriate mask for annotation from the four polygon masks generated by the pseudo-labels.

\textbf{Step 4: Manual Correction.} Since the pseudo-labels generated in Step 3 may still contain errors, annotators iteratively refine these labels until they meet our quality standards. A total of seven people were involved in labeling the dataset: three were researchers specializing in camouflage object detection (including both video and still image camouflage detection), while the remaining four were not. In this step, annotators perform two main tasks: 1) Compare multiple frames before and after the current frame to identify inaccuracies or errors in labeling, such as missing objects, incorrect object boundaries, background incorrectly labeled as the object, or newly appearing object that have not been labeled in cases with multiple camouflage objects. 2) Manually correct these errors.

\subsection{Dataset Statistics and Characteristics}

\begin{figure}
    \centering
    \includegraphics[width=1\linewidth]{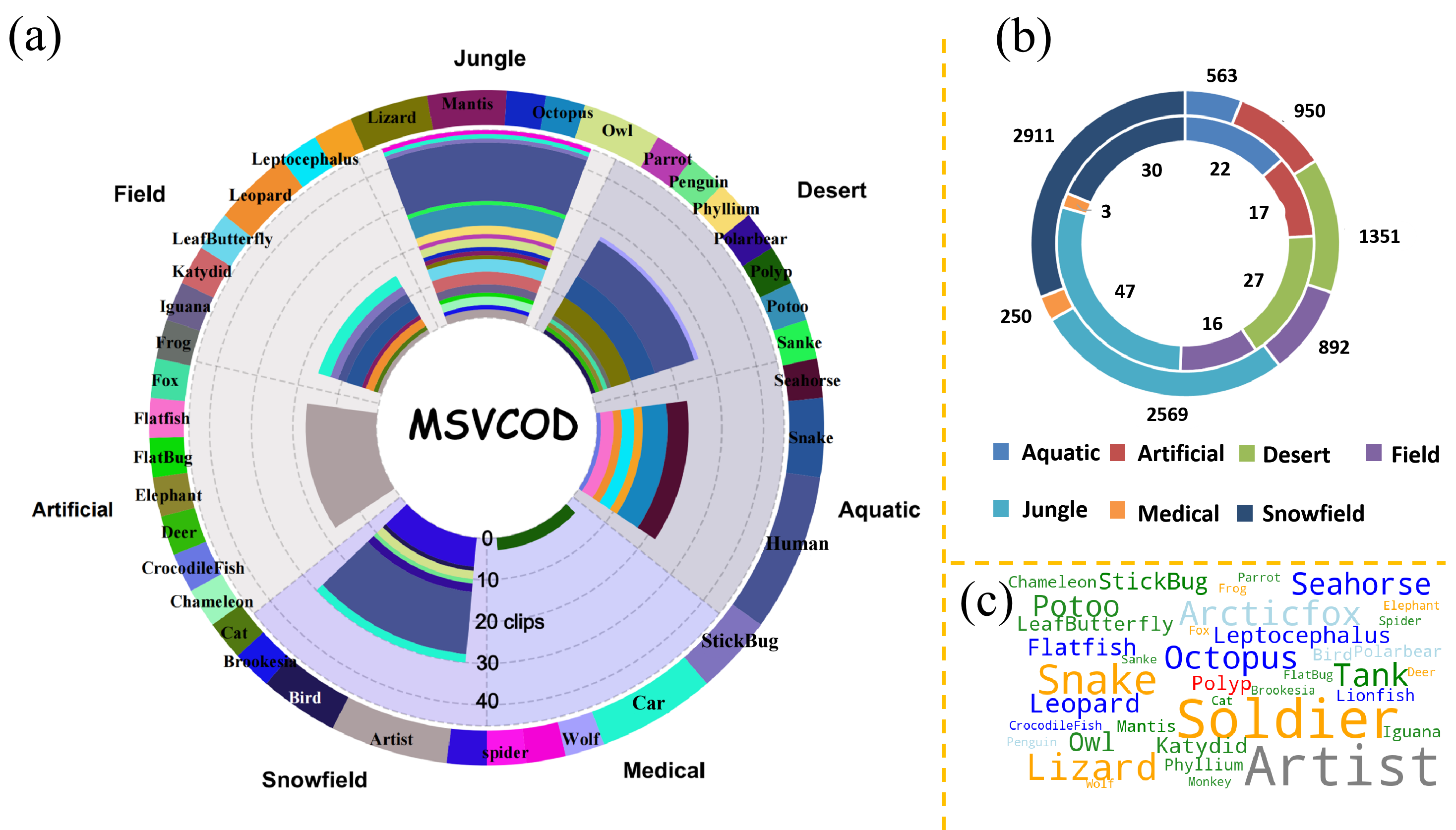}
    \caption{Statistical charts of the MSVCOD. Figure (a) represents the percentage of occurrence of the 4 types of targets in each scenario.The inner circle of (b) indicates the number of clips for the seven scenes, and the outer circle indicates the number of frames in each scene. (c) shows the word cloud of MSVCOD.}
    \label{fig:statistic graphs}
\vspace{-0.4cm}
\end{figure}





As shown in Figure \ref{fig:fig1}, the video clips cover seven distinct scenes, each representing different environmental conditions: Aquatic, depicting underwater environments with various aquatic organisms; Field, showcasing terrestrial landscapes such as rocky terrains or grasslands with sparse vegetation; Desert, featuring arid, sandy environments with minimal plant life; Jungle, characterized by dense, tropical forests with abundant foliage; Snowfield, highlighting icy, snow-covered terrains; Medical, focusing on clinical settings; and Artificial, which includes various man-made environments, such as urban areas and indoor spaces, where individuals attempt to camouflage within diverse artificial surroundings.

As shown in the bottom part of Figure \ref{fig:fig1},  the camouflage objects are categorized into four types, representing different subjects: animal, encompassing both wild and domesticated species that blend into their environments; human, referring to individuals who employ various techniques to remain concealed in different settings; medical target, such as objects requiring segmentation in clinical videos; and vehicle, involving transportation means like cars designed to evade detection in their respective environments.



Figure \ref{fig:statistic graphs} presents the video-level statistics for MSVCOD, which consists of 162 video clips. Each clip has an average duration ranging from 3 to 40 seconds and a frame rate of 6 FPS, with an average of 59 frames per video. The dataset includes bounding box, mask, instance, and category annotations, and is split into a training set (121 clips, 75\%) and a test set (41 clips, 25\%). The attribute distribution of MSVCOD is shown in Figure \ref{fig:statistic graphs}.

As demonstrated in Table \ref{tab_comparison}, compared to previous datasets, our dataset offers a richer variety of target categories, more detailed annotations. In terms of data size, although MoCA is similar in size to our dataset (141 clips vs. 162, 7617 frames vs. 9486), it only provides bounding box annotations and cannot be used for pixel-level segmentation tasks. MoCA-Mask, an improved version of MoCA, includes some pixel-level manual annotations; however, it has fewer clips and frames, with our dataset nearly doubling its size (162 clips vs. 87, 9486 frames vs. 5750).

\section{Benchmark Performance}

In this section, we present a one-stream VCOD framework that overcomes the multi-stage complexity of previous models and outperforms the current state-of-the-art video camouflage object detection models by utilizing only two frames of short-term motion features. Additionally, we provide a comprehensive evaluation of existing VCOD models.

\subsection{One-stream VCOD}

\begin{figure}
    \centering
    \includegraphics[width=1\linewidth]{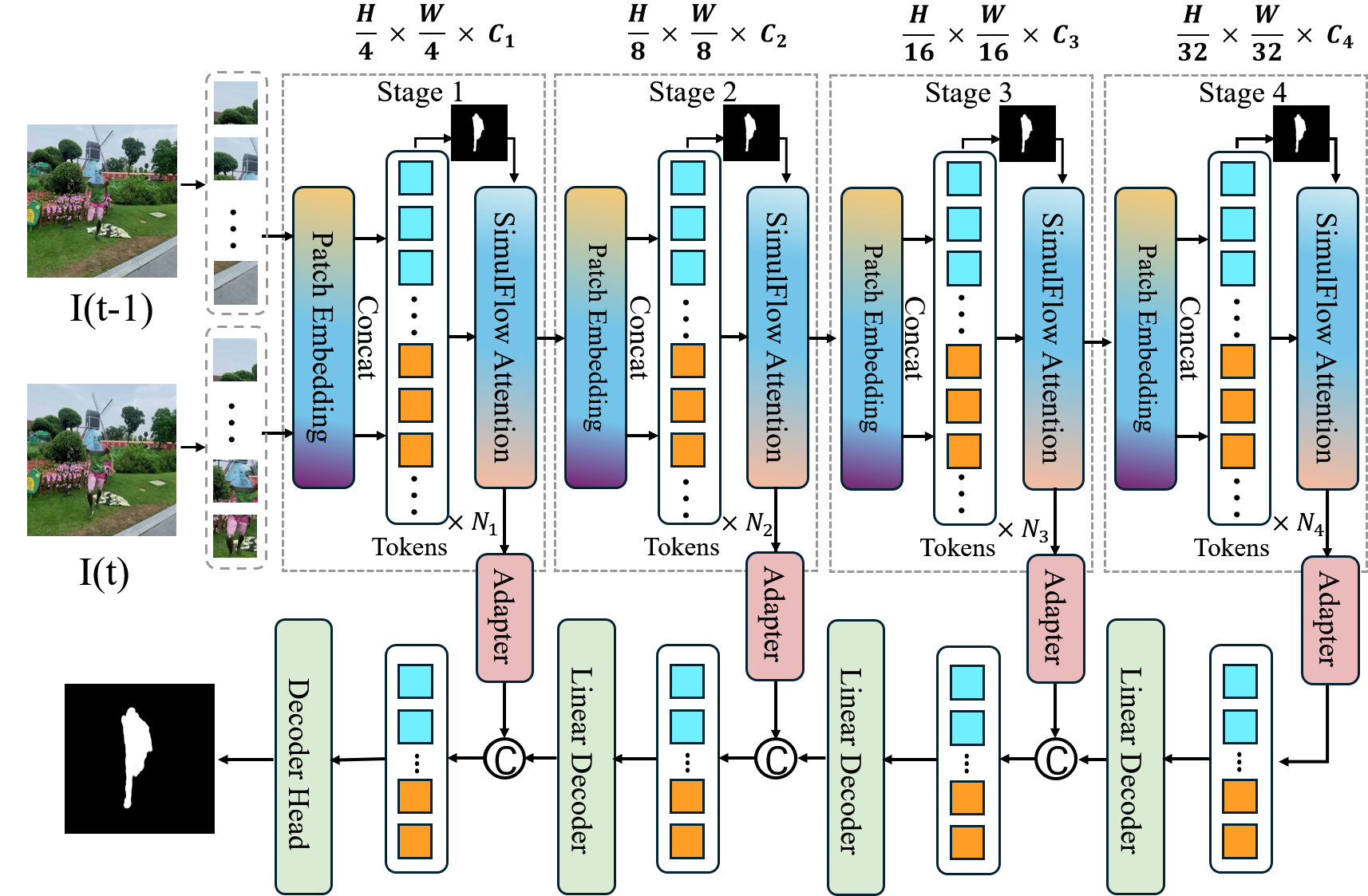}
    \caption{ Illustration of the our one-stream model pipline.}
    \label{fig:method pipline}
\vspace{-0.3cm}
\end{figure}

\textbf{Pipline overview.} 
Inspired by the one-stream architecture of tracking \cite{cui2022mixformer} and video object segmentation \cite{hong2023simulflow}, we provide a novel One-Stream video Camouflage object detection network, called OSCNet, which processes the current and previous frames simultaneously, handling motion implicitly within a one-stream architecture. The framework comprises an encoder for concurrent feature extraction and object recognition, and a fully-connected UNet-like decoder, using linear adapter layer, as illustrated in Figure \ref{fig:method pipline}. Initially, we utilize a four-stage asymmetrical transformer as the feature extraction backbone, each stage incorporating a patch embedding layer and $N_{i}$ asymmetrical attention layers. Then the tokens at four level are fed correspondingly into the Unet-like decoder to predict the camouflage object in current frame. Specifically, given the current frame image $I_{t}$ and the previous frame image $I_{t-1}$ with a resolution of $H\times  W\times 3$, we first divide them into patches of size $4\times 4$ and use a patch embedding layer to project these patches into the feature space. These patches are then input into an asymmetrical transformer with a hierarchical design to obtain multi-level feature representations of appearance.

The appearance feature tokens of the $i$ stage $\left ( i\in \left \{ 1, 2, 3, 4 \right \}  \right )$ are denoted as $F_{i}$. The features $F_{1}, F_{2}, F_{3}$ and $F_{4}$ are sequentially input into the MLP decoder, and through upsampling and multiple MLP layers. Finally, the decoder head output the prediction $P_0$ of $I_{t}$. 

\textbf{Learning Strategy.} During training, we use the weighted Binary Cross Entropy (BCE) Loss and the weighted Intersection over union (IoU) loss to optimize the model: $\mathcal{L}_{all}= \sum_{j=0}^{4}(\mathcal{L}_{bce}^{w}(P_{j},G)+\mathcal{L}_{iou}^{w}(P_{j},G))$,
where $G$ is the ground truth label, and $P_j$ are the predicted maps. When $j=1,2,3,4$, $P_j$ denotes the auxiliary loss for the four stages in Figure \ref{fig:method pipline}. $P_0$ denotes the final prediction.





\subsection{Experiments}

\begin{table*}[!ht]
    \centering
    \caption{Quantitative comparisons with state-of-the-art methods on MoCA-Mask and CAD, "$\downarrow$"/"$\uparrow$" indicates that smaller/larger is better. }
    \label{tab:comparison moca-mask}

    \resizebox{\textwidth}{!}{
    \begin{tabular}{c c ccccc c ccccc}
        \toprule
         \multirow{2}{*}{\textbf{Model}} & \multirow{2}{*}{\textbf{Input}} & \multicolumn{5}{c}{\textbf{MoCA-Mask}} & \multirow{2}{*}{} & \multicolumn{5}{c}{\textbf{CAD}} \\
        \cline{3-7} \cline{9-13}
           & & $\mathbf{S_\alpha} \uparrow$ & $\mathbf{F^{w}_{\beta}} \uparrow$ & $\mathbf{M} \downarrow$ & \textbf{mDice} $\uparrow$ & \textbf{mIoU} $\uparrow$ & & $\mathbf{S_\alpha} \uparrow$ & $\mathbf{F^{w}_{\beta}} \uparrow$ & $\mathbf{M} \downarrow$ & \textbf{mDice} $\uparrow$ & \textbf{mIoU} $\uparrow$ \\
        \cline{1-7} \cline{9-13}
         SINet(CVPR'20)\cite{fan2020camouflaged} &Image & 0.598 & 0.231 & 0.028 & 0.276 & 0.202 & & 0.636 & 0.346 & 0.041 & 0.381 & 0.283 \\
         SINet-V2(TPAMI'21)\cite{FanDP2021-ConcealedOD} &Image & 0.588 & 0.204 & 0.031 & 0.245 & 0.180 & & 0.653 & 0.382 & 0.039 & 0.413 & 0.318 \\
         ZoomNet(CVPR'22)\cite{pang2022zoom} &Image & 0.582 & 0.211 & 0.033 & 0.224 & 0.167 & & 0.633 & 0.349 & 0.033 & 0.349 & 0.273 \\
         DGNet(MIR'23)\cite{ji2023deep} &Image & 0.581 & 0.184 & 0.024 & 0.222 & 0.156 & & 0.686 & 0.416 & 0.037 & 0.456 & 0.340 \\
         FSPNet(CVPR'23)\cite{huang2023feature} &Image & 0.594 & 0.182 & 0.044 & 0.238 & 0.167 & & 0.681 & 0.401 & {0.044} & 0.238 & 0.167 \\ 
         FEDER(CVPR'23)\cite{He2023-FEDER} &Image & 0.560 & 0.165 & 0.033 & 0.194 & 0.137 & & 0.691 & 0.444 & {0.029} & 0.474 & 0.375 \\
         HitNet (AAAI'23)\cite{hu2023high} &Image & 0.623 & 0.299 & 0.019 & 0.318 & 0.254 & & 0.685 & 0.463 & 0.031 & 0.389 & 0.375 \\
        \cline{1-7} \cline{9-13}
        RCRNet(ICCV'19)\cite{yan2019semi} &Video & 0.555 & 0.138 & 0.033 & 0.171 & 0.116 & & 0.627 & 0.287 & 0.048 & 0.380 & 0.229 \\
         PNS-Net (MICCAI'21)\cite{ji2021progressively} &Video & 0.544 & 0.093 & 0.036 & 0.195 & 0.101 & & 0.655 & 0.334 & 0.032 & 0.390 & 0.290 \\
         MG (ICCV'21)\cite{yang2021self} &Video & 0.530 & 0.168 & 0.067 & 0.181 & 0.127 & & 0.606 & 0.203 & 0.059 & 0.310 & 0.176 \\
         SLT-Net (CVPR'22)\cite{cheng2022implicit} &Video  & 0.634 & 0.317 & 0.027 & 0.356 & 0.271 & & \textbf{0.696} & {0.471} & 0.031 & 0.480 & {0.392} \\
         SLT-Net-Long (CVPR'22)\cite{cheng2022implicit} &Video  & {0.631} & {0.311} & {0.026} & {0.367} & {0.279} & & {0.691} & {0.481} & \textbf{0.030} & {0.493} &{0.401} \\
         IMEX(TMM'24)\cite{hui2024implicit}  &Video & {0.661}  & {0.371} & {0.020} &{0.409} & {0.319} & &{0.695} & \textbf{0.490} & \textbf{0.030} & {0.501} & \textbf{0.412} \\
         \textbf{Ours(OSCNet)}  &Video  & \textbf{0.706} &\textbf{0.455} & \textbf{0.011} &\textbf{0.495} & \textbf{0.404} & & {0.682} & {0.476} & {0.035} & \textbf{0.513} & {0.409} \\
        \bottomrule
    \end{tabular}
    }
\end{table*}

\begin{table}[!ht]
    \centering
    \caption{Quantitative comparisons with state-of-the-art methods on MSVCOD datasets. 
    }
    \label{tab:comparison msvcod}

    \resizebox{0.47\textwidth}{!}{
    \begin{tabular}{c c ccccc}
        \toprule
         \multirow{2}{*}{\textbf{Model}} & \multirow{2}{*}{\textbf{Input}} & \multicolumn{5}{c}{\textbf{MSVCOD}} \\
        \cline{3-7}
           & &$\mathbf{S_\alpha}$ $\uparrow$ & $\mathbf{F^{w}_{\beta}}$ $\uparrow$ & $\mathbf{M}$ $\downarrow$ & \textbf{mDice} $\uparrow$ & \textbf{mIoU} $\uparrow$ \\
        \hline
         SINet\cite{fan2020camouflaged} &Image& 0.750 & 0.541 & 0.045 & 0.587 & 0.477 \\
         SINet-V2\cite{FanDP2021-ConcealedOD} &Image& 0.804 & 0.635 & 0.036 & {0.688} & 0.584 \\
         ZoomNet\cite{pang2022zoom} &Image& {0.810} & {0.656} & {0.028} & 0.684 & {0.604} \\
        \hline

        PNS-Net\cite{ji2021progressively}	&Video &0.540	&0.147	&0.117	&0.181	&0.118 \\

        MG\cite{yang2021self}	&Video &0.431	&0.219	&0.439	&0.350	&0.246  \\
        
         SLT-Net\cite{cheng2022implicit}  &Video & {0.841} & {0.716} & {0.029} & {0.766} & {0.673} \\
        \hline
         \textbf{Ours(OSCNet)} &Video & \textbf{0.845} & \textbf{0.744} & \textbf{0.026} & \textbf{0.771} & \textbf{0.695} \\
        \bottomrule
    \end{tabular}
    }
\vspace{-0.3cm}
\end{table}

\subsubsection{Implementation Details}

All input images were random cropping to produce 512x512 patches. The patches were then augmented using random flips and photometric distortion. The batch size was set to 8. The entire model was optimized using the AdamW \cite{Loshchilov2017DecoupledWD} optimizer. In terms of the training strategy, and to ensure fairness, we followed a two-stage model training approach, similar to Cheng \emph {et al.} \cite{cheng2022implicit}. In the first stage, the backbone was trained on the COD10k dataset, followed by fine-tuning on either the MoCA-mask or MSVCOD training set. Evaluation was then conducted on three video camouflage object detection datasets: CAD, MoCA-mask, and MSVCOD. During the pre-training phase, the learning rate was set to 1e-4 and decayed to 0 after 80,000 iterations. For fine-tuning, the learning rate was set to 1e-5, with a maximum of 8,000 training iterations.

\textbf{Evaluation Metrics.} We use five metrics for evaluation: S-measure ($S_\alpha$) \cite{fan2017structure} for structural similarity, F-measure with weights ($F_w^{\beta}$) \cite{margolin2014evaluate} for precision and recall, Mean Absolute Error (MAE) for pixel-level differences, Average Dice (mDice) for data similarity, and Average IoU (mIoU) for mask overlap. 

\subsubsection{Baseline Evalution}

To assess our dataset and model, we compared our model with existing state-of-the-art methods. For fairness, the performance in Table \ref{tab:comparison moca-mask} was either evaluated using the authors' published prediction images or tested using the authors' open-source weights. For testing on MSVCOD, we use the authors' open source code for training, and testing. The Table \ref{tab:comparison moca-mask} uses the follow setting: methods trained on MoCA-Mask training dataset, comparison on our MoCA-Mask and CAD test dataset; Table \ref{tab:comparison msvcod} uses the follow setting: methods trained on our MSVCOD training dataset, comparison on our MSVCOD test dataset.




From the results of Table \ref{tab:comparison moca-mask} and Table \ref{tab:comparison msvcod}, it is clear that our one-stream model performs well. Unlike previous methods that extract features from adjacent frames or image and optical flow separately and then fuse them, our model simultaneously extracts image features and incorporates motion information. Self-attention handles image feature extraction, while cross-attention uses motion features from adjacent frames. This alternating process of interactive extraction and fusion optimizes feature extraction for camouflaged objects, setting our approach apart from previous methods.

Table \ref{tab:comparison moca-mask} shows that our one-stream model consistently outperforms the previous SoTA model on both MoCA-Mask datasets. MoCA-Mask contains 87 video clips 5750 image frames (CAD only contains 9 clips, 191 frames), and our model exhibits consistent performance improvement in MoCA-Mask compared to IMEX:2.6\% improvement in mIoU, 1.7\% improvement in mDice, 1.1\% improvement in $S_\alpha$ 1.3\% improvement in $F^{w}_{\beta}$ , and 0.7\% reduction in MAE. 

Similar to SLT-Net \cite{cheng2022implicit} and IMEX \cite{hui2024implicit}, we also compared RCRNet \cite{yan2019semi}, PNS-Net \cite{ji2021progressively}, and MG \cite{yang2021self} in Table \ref{tab:comparison moca-mask}. PNS-Net, RCRNet, and MG are not methods specialized for video camouflage target detection models. PNS-Net is a model for medical polyp segmentation, while RCRNet is a semi-supervised video salient object segmentation method using only pseudo-labels, and MG is a self-supervised video object segmentation method using only optical flow. So their performance is lower than our model, SLT-Net and IMEX. To explore future possibilities on multiple tasks, we also trained and tested PNS-Net, the MG on our MSVCOD in Table \ref{tab:comparison msvcod}.


The results in Table \ref{tab:comparison msvcod} show that static image-based camouflage object detection methods (\emph {e.g.}, SINet, SINet-V2, and ZoomNet) on MSVCOD also perform well, but there is a significant gap compared to advanced VCOD methods. Since IMEX does not have available code, its performance was not evaluated. Our model outperforms SLTNET on all five metrics,
particularly with 3.9\% improvement in $F^{w}_{\beta}$, 3.2\% improvement in mIoU.

\subsubsection{Dataset Analysis.}

Comparing the test results in Exp 1 (Table \ref{tab:comparison moca-mask}) and Exp 2 (Table \ref{tab:comparison msvcod}), models generally perform worse on the MoCA-Mask test set. However, this does not mean MoCA-Mask is more complex than MSVCOD. As shown in Figure \ref{fig:msvcod_vs_moca} and \textbf{the Figure 1 and Figure 2 in the supplementary material}, the MoCA-Mask test set contains 16 clips, all with small objects (mostly between 0 and 0.04 in size). Whereas in the training set, the distribution of large and small targets is more uniform, which may be the reason why most models perform poorly on the test set of MoCA-Mask. Small objects, especially camouflaged ones, are inherently harder to detect. In contrast, the MSVCOD test set, with 41 clips, has a more diverse range of object sizes, as shown in Figures 2 and Figures 3. Thus, the MoCA-Mask test set's focus on small objects makes it more challenging, whereas MSVCOD provides a more balanced representation of object types.

\begin{figure}
    \centering
    \includegraphics[width=1\linewidth]{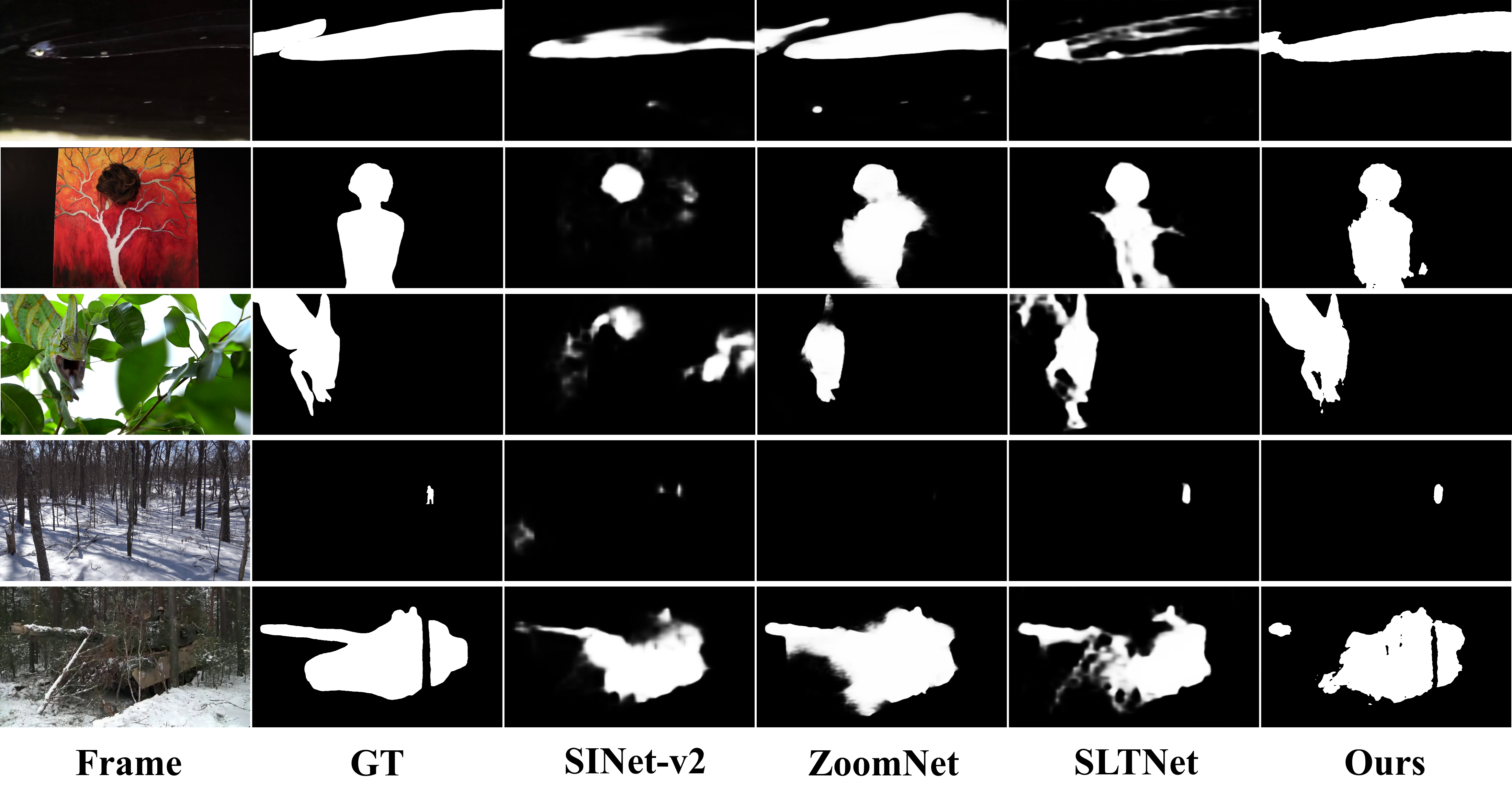}
    \caption{Visual comparisons on the MSVCOD benchmark demonstrate that our model predicts camouflaged objects more accurately across a range of challenging scenarios.}
    \label{fig:visual_comparison}
\vspace{-0.3cm}
\end{figure}

Figure \ref{fig:visual_comparison} provides a qualitative comparison of our proposed model with existing methods on representative objects (human, animal, vehicle) from MSVCOD. Our method outperforms others in both target integrity and local details (the horns of the chameleon in line 3), owing to our one-stream camouflage object detection framework.

\begin{figure}
    \centering
    \includegraphics[width=1\linewidth]{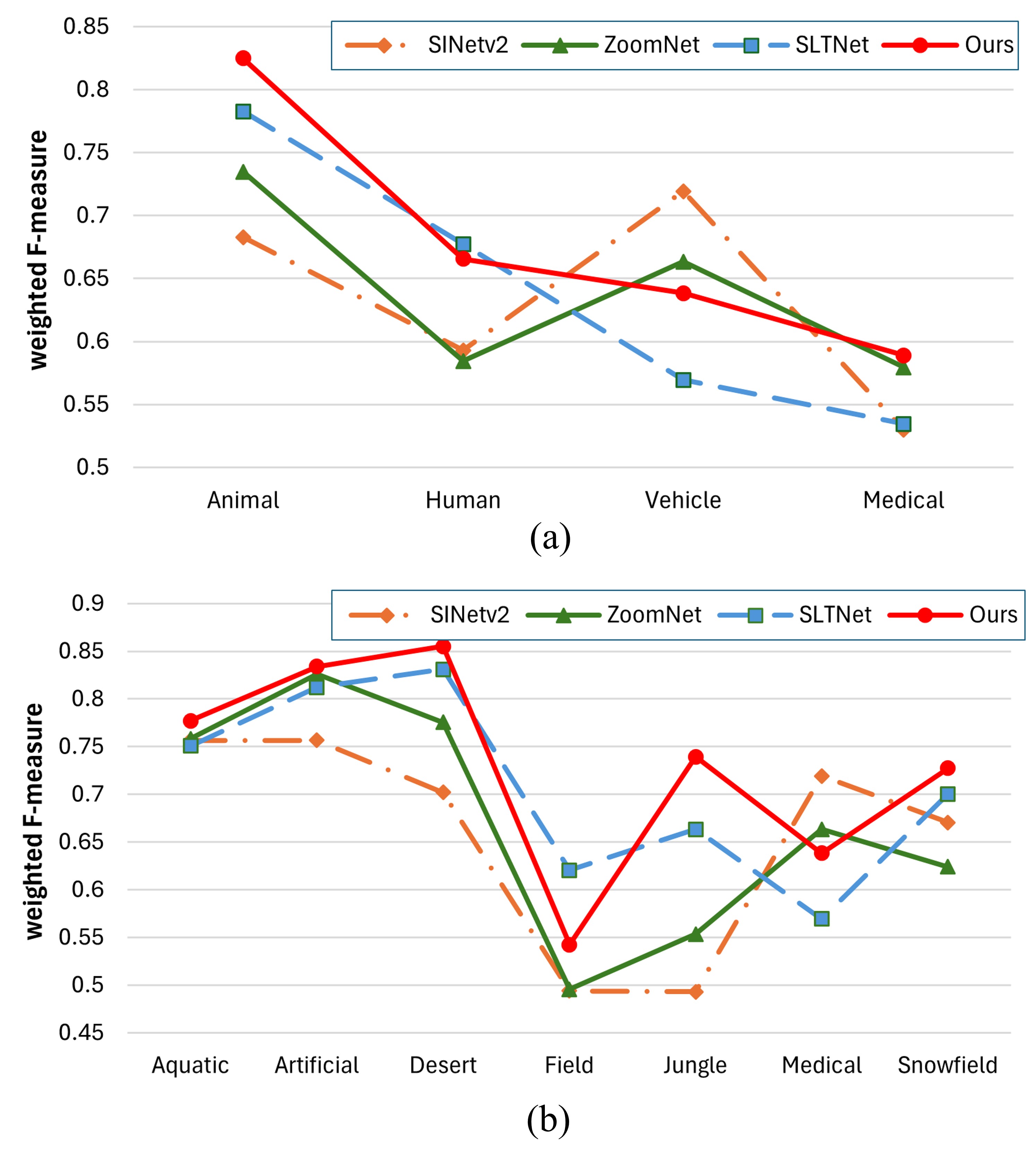}
    \caption{The results of group experiments on the MSVCOD dataset. Subgraph (a) compares the performance across major categories, and subgraph (b) contrasts the results across different scenarios. The weighted F-measure is evaluation metric.}
    \label{fig:Line chart}
\end{figure}

To analyze the model performance across four object types and seven scenarios, we plotted the line chart in Figure \ref{fig:Line chart}. In Figure \ref{fig:Line chart} (a), the performance of different models (including SINetv2, ZoomNet, and SLTNet) varies across different camouflaged objects (\emph {e.g.}, animal, human, vehicle, and medical objects). All models perform well on animals, with the best performance of $\mathbf{F^{w}_{\beta}}$ reaches 0.825, but their performance is lower for other object categories (humans, vehicles, medical), particularly for medical objects, due to the relatively small number of medical objects in MSVCOD. Figure \ref{fig:Line chart} (b) compares the model's detection performance across different scenes (Aquatic, Artificial, Desert, Field, Jungle, Medical, Snowfield). The models show significant performance differences across these scenes, which suggests that the scenarios also make a large difference to the camouflage detection model.



\begin{figure}
    \centering
    \includegraphics[width=1\linewidth]{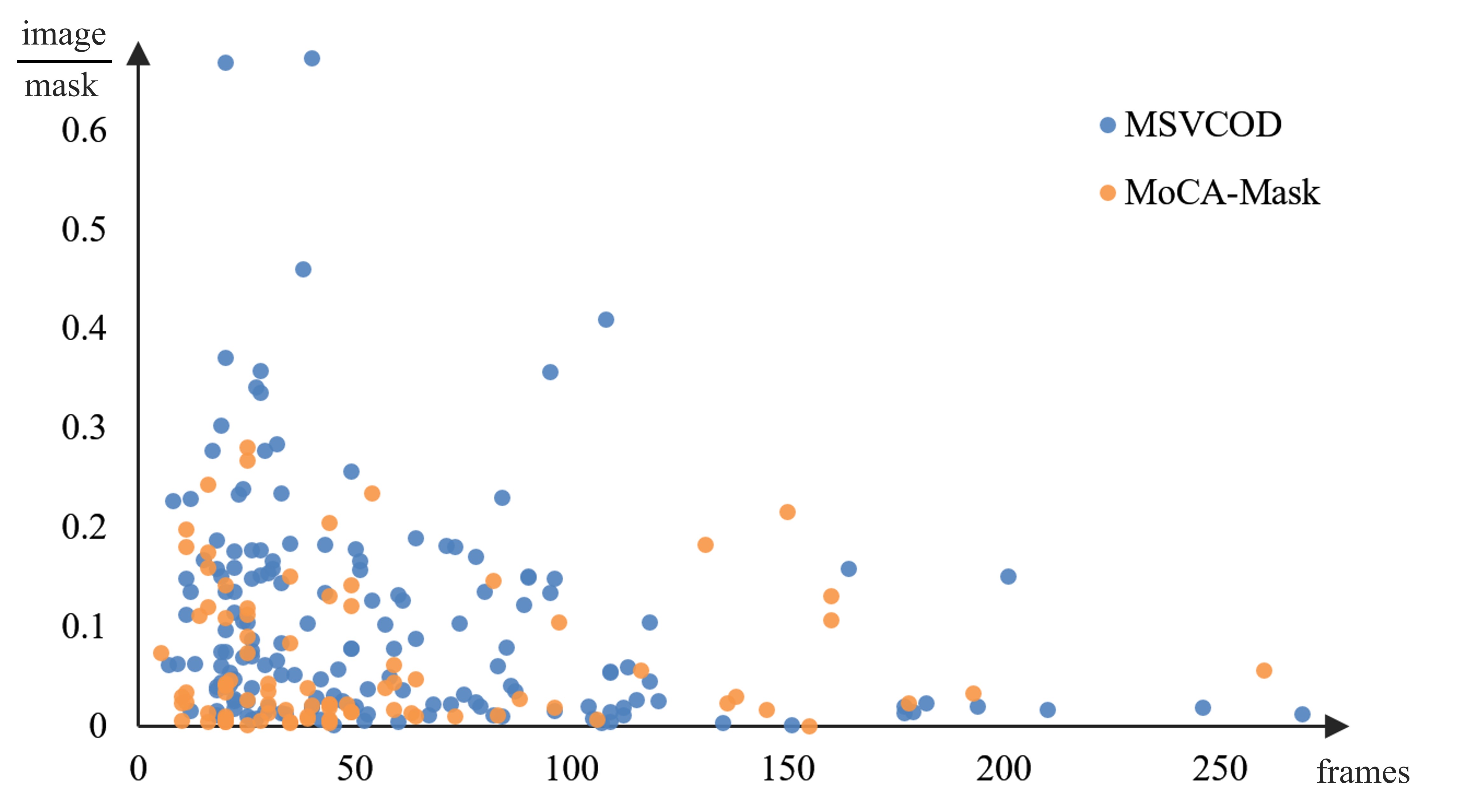}
    \caption{The scale distribution of camouflaged objects in MSVCOD and MoCA-Mask. The horizontal coordinate represents the number of frames in the clips, and the vertical coordinate represents the ratio of the camouflage object to the image size. It can be seen that our MSVCOD dataset has a much larger range of camouflage object scales.}
    \label{fig:msvcod_vs_moca}
\vspace{-0.4cm}
\end{figure}


\subsection{Limitation Discussion}
In Figure. \ref{fig:statistic graphs}, it can be seen that the number of medical image videos is limited, and the unbalanced distribution of data is a limitation of our dataset. On the other hand, the length of the videos is not long enough (only 10-15 seconds on average) due to the limitation of available data. Although we had access to a small number of videos lasting longer than 1 minute, only a very small portion of them were available. We had to intercept some of them to keep close to other parts of the video clips. Limited by the videos we could collect, the number of camouflage instances in our videos is relatively small, with most video clips having only one camouflage object.

\section{Conclusion}


This paper introduces the MSVCOD dataset, a novel video camouflage target detection dataset featuring multiple scenes and target categories. We design a semi-automatic iterative annotation pipeline to efficiently annotate this large-scale dataset, which can serve as a reference for similar tasks. The dataset includes 162 video clips across seven scenarios (underwater, land, desert, jungle, snowfield, medical, and art) and four object categories (animal, human, medical, and vehicle). It provides bounding box, mask, instance, and category annotations, divided into training and test sets. We also propose a one-stream model for video camouflage object detection, achieving state-of-the-art performance on multiple datasets. Experimental results show that MSVCOD facilitates the extension of video camouflage object detection to diverse scenes and object types.

{
    \small
    \bibliographystyle{ieeenat_fullname}
    \bibliography{main}
}


\end{document}